\documentclass[10pt]{article}
\usepackage[margin=1in]{geometry}

\usepackage[utf8]{inputenc}
\usepackage{amsmath}
\usepackage{amssymb}
\usepackage{mathtools}
\usepackage{csquotes}
\usepackage[normalem]{ulem}

\usepackage{authblk}
\usepackage{graphicx}
\usepackage[font=footnotesize]{caption}
\usepackage[ruled,vlined]{algorithm2e}
\usepackage{afterpage}
\usepackage{wrapfig}

\newcommand{\defeq}{\vcentcolon=}

\newcommand{\argminF}{\mathop{\mathrm{argmin}}\limits}
\newcommand{\argmin}[1]{\arg\min_{#1} \ }

\renewcommand{\star}[1]{#1^*}

\newcommand{\R}{\mathbb{R}}
\newcommand{\N}{\mathbb{N}}
\newcommand{\from}{\colon}
\renewcommand{\to}{\rightarrow}

\renewcommand{\epsilon}{\varepsilon}

\newcommand{\grad}[2]{\nabla #1\left(#2\right)}
\newcommand{\hess}[2]{\nabla^2 #1\left(#2\right)}
\newcommand{\norm}[1]{\left\|#1\right\|}
\newcommand{\sumsq}[1]{\norm{#1}^2}

\newcommand{\eqnref}[1]{Eqn.~\ref{#1}}
\newcommand{\figref}[1]{Fig.~\ref{#1}}
\newcommand{\secref}[1]{Sec.~\ref{#1}}
\newcommand{\appref}[1]{Appendix~\ref{#1}}
\newcommand{\algoref}[1]{Algorithm~\ref{#1}}

\newcommand{\successcolor}{blue}
\newcommand{\failcolor}{orange}

\newcommand{\equalcontrib}{%
  \textsuperscript{$\dagger$}%
}
\title{Critical Point-Finding Methods Reveal
Gradient-Flat Regions of Deep Network Losses}
\author[1,2]{\footnote{To whom correspondence should be addressed:
charlesfrye@berkeley.edu}%
Charles G. Frye}
\author[1,3]{James Simon}
\author[1,4]{Neha S. Wadia}
\author[1,4]{Andrew Ligeralde}
\author[1,2,3,4]{\equalcontrib Michael R. DeWeese}
\author[1,2,5,6]{\footnote{These authors contributed equally to the work.}%
Kristofer E. Bouchard}
\affil[1]{Redwood Center for Theoretical Neuroscience,
University of California, Berkeley, CA, USA}
\affil[2]{Helen Wills Neuroscience Institute,
University of California, Berkeley, CA, USA}
\affil[3]{Department of Physics,
University of California, Berkeley, CA, USA}
\affil[4]{Biophysics Graduate Group,
University of California, Berkeley, CA, USA}
\affil[5]{Biological Systems and Engineering Division,
Lawrence Berkeley National Lab, Berkeley, CA, USA}
\affil[6]{Computational Research Division,
Lawrence Berkeley National Lab, Berkeley, CA, USA}
\date{}

\begin{document}

\maketitle
\vspace*{-3em}
\begin{abstract}

Despite the fact that the loss functions of
deep neural networks are highly non-convex,
gradient-based optimization algorithms converge to approximately the same
performance from many random initial points.
One thread of work has focused on explaining this phenomenon
by characterizing the local curvature near critical points
of the loss function, where the gradients are near zero,
and demonstrating that neural network losses enjoy a
no-bad-local-minima property and an abundance of saddle points.
We report here that the methods used to find these putative critical points
suffer from a bad local minima problem of their own:
they often converge to or pass through regions where
the gradient norm has a stationary point.
We call these \emph{gradient-flat regions},
since they arise when the gradient is
approximately in the kernel of the Hessian,
such that the loss is locally approximately linear, or flat,
in the direction of the gradient.
We describe how the presence of these regions necessitates care
in both interpreting past results that claimed to find
critical points of neural network losses
and in designing second-order methods for optimizing neural networks.

\end{abstract}

\section{Introduction}

Large neural networks are surprisingly easy to optimize~\cite{sun2019},
despite the substantial non-convexity of
the loss as a function of the parameters~\cite{goodfellow2015}.
In particular,
it is usually found that changing the random initialization
has no effect on performance,
even though it can change the model learned
by gradient-based optimization methods%
~\cite{garipov2018}.
Understanding the cause of trainability from random initial conditions
is critical for the development of new architectures and optimization methods,
which must otherwise just hope to retain this favorable property
based on heuristics.

One possible explanation for this phenomenon
is based on the stationary points of gradient-based
optimization methods.
These methods are stationary
when the gradient of the loss function is 0,
at the \emph{critical points} of the loss.
Critical points are classified by their Morse index,
or the degree of local negative curvature
(i.e., the relative number of dimensions in parameter space
along which the curvature is negative).
Since, among all critical points,
gradient descent methods only converge to those points with index 0%
~\cite{lee2016},
which includes local minima, it has been argued that
large neural networks are easy to train because
their loss functions for many problems
only have local minima at values of the loss
close to or at the global optimum.
This is known as 
the \enquote{no-bad-local-minima} property.
Previous work~\cite{dauphin2014,pennington2017} has
reported numerical evidence for a convex relationship between index and loss
that supports the hypothesis that neural network loss functions
have the no-bad-local-minima property:
for low values of the loss, only low values of the index were observed,
whereas for high values of the loss, only high values of the index were observed.
However, more recent theoretical work
has indicated that there are in fact
bad local minima on neural network losses
in almost all cases%
~\cite{ding2019}.

The validity of the numerical results
depends on the validity of
the critical point-finding algorithms, and
the second-order critical point-finding algorithms
used in~\cite{dauphin2014} and~\cite{pennington2017}
are not in fact guaranteed to find critical points
in the case where the Hessian is singular.
In this case, the second-order information used
by these critical point-finding methods becomes unreliable.

\begin{figure}[!h]
    \centering
    \includegraphics[width=0.6\linewidth]{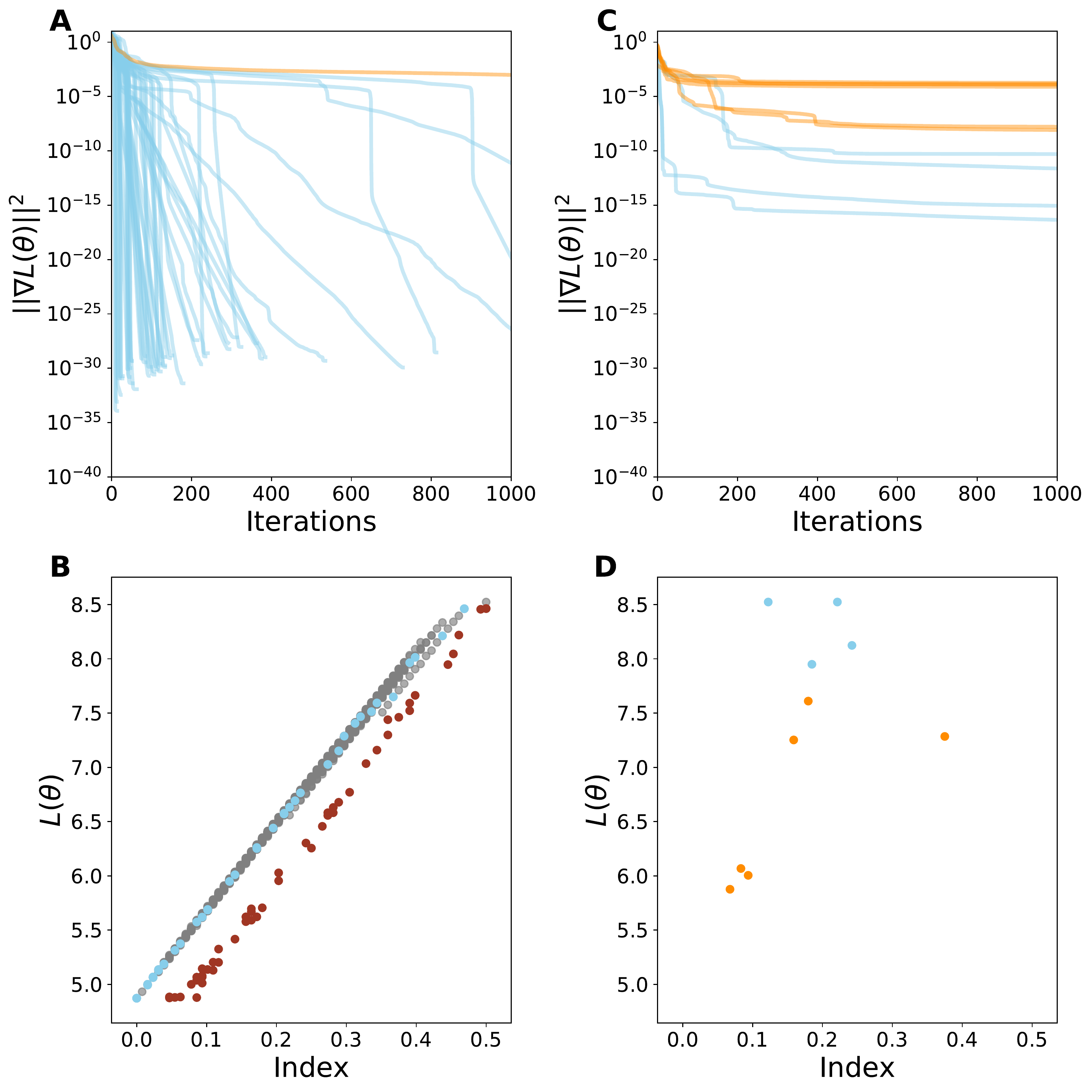}
    \caption{\textbf{Newton Methods that Find Critical Points on a Linear Network
    Fail on a Non-Linear Network}.
    \textbf{A-B}: Newton-MR on a linear autoencoder
    applied to multivariate Gaussian data,
    as in~\cite{frye2019}.
    \emph{A}:
    Squared gradient norms of the loss $L$,
    as a function of the parameters $\theta$,
    across iterations of Newton-MR,
    colored by whether, after the first of early termination or 1000 epochs,
    squared gradient norms are below 1e-10 (\successcolor{})
    or not (\failcolor{}).
    \emph{B}:
    The loss and Morse index of putative
    and actual critical points, with ground truth.
    The Morse index is defined as the fraction of negative eigenvalues.
    Analytically-derived critical points in gray,
    points from the end of runs that terminate
    below a squared gradient norm of 1e-10 in light \successcolor{},
    and points from trajectories stopped early,
    once they pass a squared gradient norm of 1e-2, in dark red.
    \textbf{C-D}:
    As in \emph{A}-\emph{B},
    on the same network architecture and data,
    but with Swish~\cite{ramachandran2017} non-linear activations
    instead of identity activations.
    \emph{D}:
    Loss and Morse index of putative critical points.
    Points with squared gradient norm above 1e-10 in \failcolor{},
    those below 1e-10 in \successcolor{}.
    Analytical expressions for critical points are not available
    for this non-linear network.}
    \label{fig:nmr_comparison}
\end{figure}

Neural network loss Hessians are typically highly singular~\cite{sagun2017},
and poor behavior of Newton-type critical point-finding methods has been reported
in the neural network case~\cite{coetzee1997},
casting doubt on the completeness and accuracy of the results
in~\cite{dauphin2014} and~\cite{pennington2017}.
\cite{frye2019} verified that second-order methods
can in fact find high-quality approximate critical points
for linear neural networks,
for which the analytical form of the critical points is known%
~\cite{baldi1989},
providing ground truth.
In particular, the two phase convergence pattern
predicted by the classical analysis of Newton methods~\cite{nocedal2006}
is evident:
a linear phase followed a short,
local supralinear phase
(\figref{fig:nmr_comparison}A).
The supralinear convergence is visible in the
\enquote{cliffs} in the \successcolor{} traces in
\figref{fig:nmr_comparison}A,
where the convergence rate suddenly improves.
With a sufficiently strict cut-off on the gradient norms,
the correct loss-index relationship obtained analytically
(\figref{fig:nmr_comparison}B, gray points)
is shared by the points obtained numerically
(\figref{fig:nmr_comparison}B,
light \successcolor{} points).
With an insufficiently strict cutoff,
the loss-index relationship implied by the observed points
is far from the truth
(\figref{fig:nmr_comparison}B,
dark red points)

Unfortunately, good performance on linear networks
does not guarantee good performance on non-linear networks.
When applied to a non-linear network,
even with the same data,
the behavior of these Newton methods changes dramatically
for the worse
(\figref{fig:nmr_comparison}C).
No runs exhibit supralinear convergence
and the gradient norms at termination are many orders of magnitude larger.
These are not the signatures of a method converging
to a critical point,
even though gradient norms are sometimes still
under the thresholds reported
in~\cite{pennington2017} and~\cite{frye2019}
(no threshold reported in~\cite{dauphin2014}).
This makes it difficult to determine whether the
putative loss-index relationship measured from these critical points
(\figref{fig:nmr_comparison}D)
accurately reflects the loss-index relationship
at the true critical points of the loss function.

In this paper,
we identify a major cause of this failure
for second-order critical point-finding methods:
\emph{gradient-flat regions},
where the gradient is approximately in the kernel of the Hessian.
In these regions, the loss function
is locally approximately linear along the direction of the gradient,
whether or not the gradient is itself small,
as would be the case near a true critical point.
We first define gradient-flatness (\secref{sec:theory})
and explain, with a low-dimensional example (\secref{sec:toy}),
why it is problematic for second-order methods:
gradient-flat points can be \enquote{bad local minima}
for the problem of finding critical points.
We then provide evidence that gradient-flat regions
are encountered when applying
the Newton-MR algorithm to a deep neural network loss
(\secref{sec:results}, \appref{app:mlp}).
Furthermore, we show that, though gradient-flat regions need not
contain actual critical points,
the loss-index relationship looks strikingly similar to that reported
in~\cite{dauphin2014} and~\cite{pennington2017},
suggesting that these previous studies may have found gradient-flat regions,
not critical points.
Finally, we note the implications of gradient-flatness for the
design of second-order methods for use in optimizing neural networks:
in the presence of gradient-flatness,
approximate second-order methods,
like K-FAC~\cite{martens2015} and Adam~\cite{kingma2014}
may be preferable to exact second-order methods even without taking computational cost into account.

\section{Gradient-Flat Points are Stationary Points for Second-Order Methods}\label{sec:gfp}

In this section,
we introduce and define gradient-flat points and
explain why they are problematic for second-order
critical point-finding methods,
with the help of a low-dimensional example
to build intuition.
In numerical settings and in high dimensions,
approximately gradient-flat points are also important,
and so we define a quantitative index of gradient-flatness
based on the residual norm of the Newton update.
Connected sets of these numerically gradient-flat points
are gradient-flat regions,
which cause trouble for 
second-order critical point-finding methods.

\subsection{At Gradient-Flat Points, the Gradient Lies in the Hessian's Kernel}\label{sec:theory}

Critical points are of interest because
they are points where the first-order approximation
of a function$f$ at a point\footnote{Note that,
for a neural network loss function,
the variable we take the gradient with respect to,
here $x$, is the vector of parameters, $\theta$,
not the data, which is often denoted with an $x$.}
$x+\delta$ based on the local information at $x$
\begin{equation}
    f(x + \delta) \approx f(x) + \grad{f}{x}^\top \delta
\end{equation}
is constant, indicating that they are the stationary points
of first-order optimization algorithms
like gradient descent and its accelerated variants.
By stationary point, we mean a point at which the proposed
updates of an optimization algorithm are zero.
Stationary points need not be
points to which the algorithm converges.
For example,
gradient descent almost surely only converges
to critical points with index 0~\cite{lee2016,jin2018a},
even though its stationary points include saddle points and maxima.
However, the curvature properties of non-minimal stationary points show up
in proofs of the convergence rates of first order optimizers,
e.g.~\cite{jin2018a,jin2018b,jin2018c},
and so knowledge of their properties can guide algorithm design.

In searching for critical points,
it is common to use a linear approximation to the behavior of the gradient
at a point $x + p$ given the local information at a point $x$
\begin{equation}
    \grad{f}{x + p} \approx \grad{f}{x} + \hess{f}{x}p
\end{equation}
Because these methods rely on a quadratic approximation of the original function $f$,
represented by the Hessian matrix of second partial derivatives,
we call them second-order critical point-finding methods.

The approximation on the right-hand side is constant whenever
$p$ is an element of $\ker{\hess{f}{x}}$,
where $\ker{M}$ is notation for the kernel of a matrix $M$
--- the subspace that $M$ maps to 0.
When $\hess{f}{x}$ is non-singular,
this is only satisfied when $p$ is $0$,
so if we can define an update rule such that
$p=0$ iff $\grad{f}{x}=0$,
then, for non-singular Hessians,
we can be sure that our method is stationary only at critical points.

In a Newton-type method,
we achieve this by selecting our step by
solving for the zeroes of this linear approximation,
i.e.~the Newton system,
$$
0 = \grad{f}{x} + \hess{f}{x}p
$$
which has solution
$$
p = -\hess{f}{x}^{+}\grad{f}{x}
$$
where the matrix $M^+$ is the Moore-Penrose
pseudoinverse of the matrix $M$,
obtained by performing the singular value decomposition,
inverting the non-zero singular values,
and recomposing the SVD matrices in reverse order.
The Newton update $p$ is zero iff $\grad{f}{x}$ is 0
for a non-singular Hessian,
for which the pseudo-inverse is simply the inverse.
For a singular Hessian,
the update $p$ is zero iff $\grad{f}{x}$ is
in the kernel of the pseudoinverse.
Note that if the Hessian is constant as a function of $x$,
the linear model of the gradient is exact and
this algorithm converges in a single step.

Within the vicinity of a critical point,
this algorithm converges extremely quickly~\cite{nocedal2006},
but the guarantee of convergence is strictly local.
Practical Newton methods in both
convex optimization~\cite{boyd2004} and
non-linear equation solving~\cite{nocedal2006,izmailov2014}
often compare multiple possible choices of $p$
and select the best one according to a
\enquote{merit function} applied to the gradients
which has a global minimum
for each critical point.
Such algorithms have broader guarantees of global convergence.
A common choice for merit function
is the squared norm,
$$
g(x) = \frac{1}{2}\sumsq{\grad{f}{x}}
$$

In gradient norm minimization~\cite{mciver1972},
we optimize this merit function directly.
The gradients of this method are
$$
\grad{g}{x} = \hess{f}{x}\grad{f}{x}
$$
As with Newton methods, in the invertible case
the updates are zero iff $\grad{f}{x}$ is 0.
In the singular case,
the updates are zero if the gradient is in the Hessian's kernel.
Because this method is framed as the minimization of a scalar function,
it is compatible with first-order optimization methods,
which are more commonly implemented and better supported
in neural network libraries.

However, neural network Hessians are generally singular,
especially in the overparameterized case~\cite{sagun2017,ghorbani2019},
meaning the kernel is non-trivial,
and so neither class of methods can guarantee
convergence to critical points.
Newton's method can diverge,
oscillate, or behave chaotically~\cite{griewank1983}.
The addition of merit function-based upgrades
can remove these behaviors,
but it cannot guarantee convergence to critical points~\cite{powell1970,griewank1983}.
The gradient norm minimization method described
in~\cite{pennington2017}
was previously proposed and this flaw pointed out
twice in the field of chemical physics ---
once in the 1970s
(proposed~\cite{mciver1972}, critiqued~\cite{cerjan1981})
and again in the 2000s
(proposed simultaneously~\cite{angelani2000,broderix2000},
critiqued~\cite{doye2002}).

What are the stationary points, besides critical points,
for these two method classes in the case of singular Hessians?
It would seem at first that they are different:
for gradient norm minimization,
when the gradient is in the Hessian's kernel;
for Newton-type methods,
when the gradient is in the Hessian's pseudoinverse's kernel.
In fact, however,
these conditions are identical, due to the Hessian's symmetry\footnote{
Indeed, the kernel of the pseudo-inverse is equal
to the kernel of transpose,
as can be seen from the singular value decomposition,
and the Hessian is equal to its transpose because it is symmetric.
See~\cite{strang1993}.},
and so both algorithms share a broad class of stationary points.

These stationary points have been identified previously,
but nomenclature is not standard:
Doye and Wales, studying gradient norm minimization,
call them \emph{non-stationary points}~\cite{doye2002},
since they are non-stationary with respect to the function $f$,
while Byrd et al., studying Newton methods,
call them \emph{stationary points}~\cite{byrd2004},
since they are stationary with respect to the merit function $g$.
To avoid confusion between these incommensurate conventions
or with the stationary points of the function $f$,
we call a point where the gradient lies in the kernel
of the Hessian a \emph{gradient-flat} point.
This name was chosen because a function is \emph{flat}
when its Hessian is 0, meaning every direction is in the kernel,
and so it is locally flat around a point in a given direction
whenever that direction is in the kernel of the Hessian at that point.
Note that, because $0 \in \ker$ for all matrices,
every critical point is also a gradient-flat point,
but the reverse is not true.
When we wish to explicitly refer to gradient-flat points
which are not critical points,
we will call them \emph{strict} gradient-flat points.
At a strict gradient-flat point, the function is,
along the direction of the gradient,
locally linear up to second order.

There is an alternative view of gradient-flat points
based on the squared gradient norm merit function.
All gradient-flat points are stationary points
of the gradient norm,
which may in principle be local minima, maxima, or saddles,
while the global minima of the gradient norm are critical points.
When they are local minima of the gradient norm,
they can be targets of convergence
for methods that use
first-order approximations of the gradient map,
as in gradient norm minimization and in Newton-type methods.
Strict gradient-flat points, then,
can be \enquote{bad local minima} of the gradient norm,
and therefore prevent the convergence of
second-order root-finding methods
to critical points,
just as bad local minima of the loss function
can prevent convergence of first-order optimization methods
to global optima.

Note that Newton methods cannot be demonstrated to converge
only to gradient-flat points~\cite{powell1970}.
Furthermore, Newton convergence can be substantially slowed
when even a small fraction of the gradient
is in the kernel~\cite{griewank1983}.
Below we will see that,
while a Newton method applied to a neural network loss
sometimes converges to and almost always encounters strict gradient-flat points,
the final iterate is not always either
a strict gradient-flat point or a critical point.

\subsection{Convergence to Gradient-Flat Points Occurs in a Low-Dimensional Quartic Example}\label{sec:toy}

\begin{figure}
    \centering
    \includegraphics[width=\linewidth]{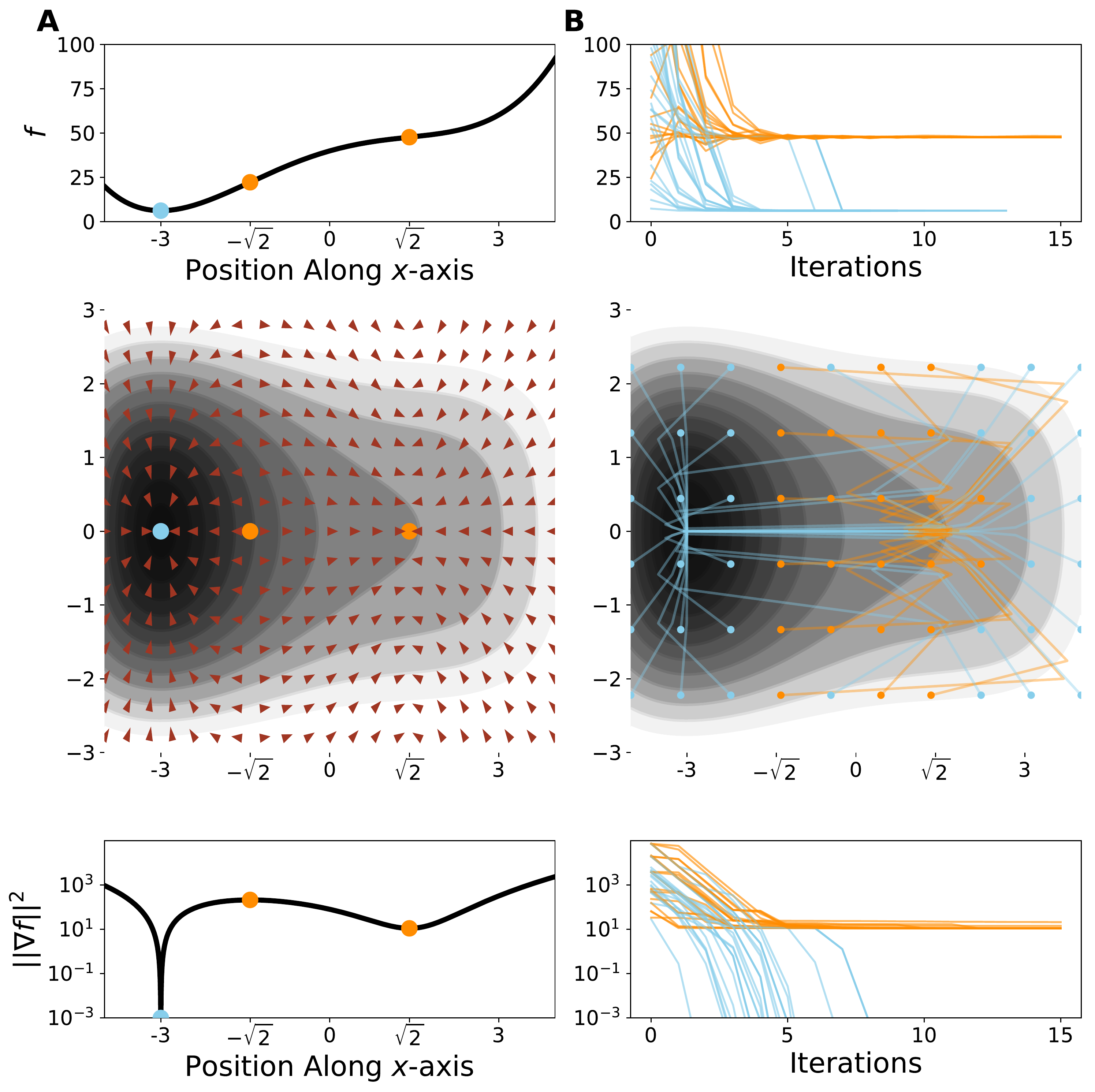}
    \caption{\textbf{Stationarity of and Convergence to
    a Strict Gradient-Flat Point on a Quartic Function}.
    \textbf{A}:
    Critical and strict gradient-flat points
    of quartic $f(x,y)$ (defined in \eqnref{eqn:toy}).
    Middle panel:
    $f(x,y)$
    plotted in color 
    black, low values; white, high values),
    along with the direction of the Newton update $p$
    as a (notably non-smooth) vector field (red).
    Stationary points of
    the squared gradient norm merit function $g$ are indicated:
    strict gradient-flat points in \failcolor{},
    the critical point in \successcolor{}.
    Top and bottom panels:
    The value (top) and squared gradient norm (bottom)
    of $f$ as a function of $x$ value
    with $y$ fixed at 0.
    The $x$ axis is shared between panels.
    \textbf{B}:
    Performance and trajectories of Newton-MR~\cite{roosta2018}
    on~\eqnref{eqn:toy}.
    Runs that terminate near a strict gradient-flat point
    are in \failcolor{},
    while those that terminate near
    a critical point are in \successcolor{}.
    Middle panel:
    Trajectories of Newton-MR laid over
    $f(x, y)$.
    $x$ and $y$ axes are shared with the middle panel of
    \emph{A}.
    Initial values indicated with scatter points.
    Top and bottom panels:
    Function values (top) and squared gradient norms (bottom)
    of Newton-MR trajectories as a function of iteration.
    The $x$ axis is shared between panels.}
    \label{fig:toy}
\end{figure}

The difficulties that gradient-flat points pose for Newton methods
can be demonstrated with a polynomial example in two dimensions,
plotted in \figref{fig:toy}A.
Below,
we will characterize
the strict gradient-flat (\failcolor{})
and critical (\successcolor{}) points of this function
(\figref{fig:toy}A).
Then, we will observe the behavior of a practical Newton
method applied to it (\figref{fig:toy}B-C)
and note similarities to the results in \figref{fig:nmr_comparison}.
We will use this simple, low-dimensional example
to demonstrate principles useful
for understanding the results of applying
second-order critical point-finding methods to more complex,
higher-dimensional neural network losses.

As our model function, we choose
\begin{equation}\label{eqn:toy}
    f(x, y) = 1/4 x^4 - 3x^2 + 9x + 0.9y^4 + 5y^2 + 40
\end{equation}

It is plotted in~\figref{fig:toy}A, middle panel.
This quartic function has two affine subspaces
of points with non-trivial Hessian kernel,
defined by $[\pm\sqrt{2}, y]$.
The kernel points along the $x$ direction and so
is orthogonal to this affine subspace at every point.
As a function of $y$, $f$ is convex,
with one-dimensional minimizers at $y=0$.
The strict gradient-flat points occur at the intersections
of these two sets:
one strict gradient-flat point at $[\sqrt{2}, 0]$,
which is a local minimum of the gradient norm,
and one at $[-\sqrt{2}, 0]$,
which is a saddle of the same
(\figref{fig:toy}A, \failcolor{} points, all panels).
In the vicinity of these points, the gradient is,
to first order, constant along the $x$-axis,
and so the function is locally linear or flat.
These points are gradient-flat but 
neither is a critical point of $f$.
The only critical point is located at the minimum of the polynomial,
at $[-3, 0]$
(\figref{fig:toy}A, \successcolor{} point, all panels),
which is also a global minimum of the gradient norm.
The affine subspace that passes through
$[-\sqrt{2}, 0]$ divides the space into two
basins of attraction, loosely defined,
for second-order methods:
one, with initial $x$-coordinate $x_0<-\sqrt{2}$,
for the critical point of $f$
and the other for the strict gradient-flat point.
Note that the vector field in the middle panel shows update directions
for the pure Newton method,
which can behave extremely poorly in the vicinity
of singularities~\cite{powell1970,griewank1983},
often oscillating and converging very slowly
or diverging.

Practical Newton methods use techniques like
damping and line search to improve behavior~\cite{izmailov2014}.
To determine how a practical Newton method
behaves on this function,
we focus on the case of Newton-MR~\cite{roosta2018},
which uses the MR-QLP~\cite{choi2011} solver%
\footnote{MR-QLP, short for MINRES-QLP,
is a Krylov subspace method akin to conjugate gradient
but specialized to the symmetric, indefinite and ill-conditioned case,
which makes it well-suited to this problem and to neural network losses.}
to compute the Newton update
and back-tracking line search
with the squared gradient norm merit function
to select the step size.
Pseudocode for this algorithm is provided in
~\appref{app:nmr}.
This method was found to perform better than
a damped Newton method
and gradient norm minimization
on finding the critical points of a linear autoencoder in~\cite{frye2019}.
Results are qualitatively similar for damped Newton methods
with a squared gradient norm merit function.

The results of applying Newton-MR
to~\eqnref{eqn:toy} are shown in%
~\figref{fig:toy}B.
The gradient-flat point is attracting
for some trajectories
(\failcolor{}),
while the critical point is attracting for others
(\successcolor{}).
For trajectories that approach the strict gradient-flat point,
the gradient norm does not converge to 0,
but converges to a non-zero value near 10
(\failcolor{} trajectories; \figref{fig:toy}B, bottom panel).
This value is typically several orders of magnitude lower
than the initial point, and so would appear to be close to 0
on a linear scale that includes the gradient norm of the initial point.
Since log-scaling of loss functions is uncommon in machine learning,
as losses do not always have minima at 0,
second-order methods apporaching gradient-flat points
can appear to converge to critical points
if typical methods for visually assessing convergence are used.

There are two interesting and atypical behaviors worth noting.
First, the trajectories tend to oscillate
in the vicinity of the gradient-flat point
and converge more slowly
(\figref{fig:toy}B, middle panel, \failcolor{} lines).
Updates from points close to the affine subspace where the Hessian has a kernel,
and so which have an approximate kernel themselves,
sometimes jump to points where the Hessian doesn't have an approximate kernel.
This suggests that, when converging towards a gradient-flat point,
the degree of flatness will change iteration by iteration.
Second, some trajectories begin in the nominal basin
of attraction of the gradient-flat point
but converge to the critical point
(\figref{fig:toy}B, middle panel, \successcolor{} points
with $x$-coordinate $>-\sqrt{2}$).
This is because the combination of back-tracking line search
and large proposed step sizes means that occasionally,
very large steps can be taken, based on non-local features of the function.
Indeed,
back-tracking line search is a limited form of global optimization
and the ability of line searches
to change convergence behaviors predicted from local properties
on nonconvex problems
is known~\cite{nocedal2006}.
Since the back-tracking line search is based on the gradient norm,
the basin of attraction for the true critical point,
which has a lower gradient norm than the gradient-flat point,
is much enlarged relative to that for the gradient-flat point.
This suggests that Newton methods
using the gradient norm merit function will be biased towards finding gradient-flat points
that also have low gradient norm.

\subsection{Approximate Gradient-Flat Points and Gradient-Flat Regions}\label{sec:approxgfp}
Analytical arguments focus on exactly gradient-flat points,
where the Hessian has an exact kernel
and the gradient is entirely within it.
In numerical settings,
it is almost certain no matrix will have an exact kernel,
due to rounding error.
For the same reason, the computed gradient vector will generically not lie entirely
within the exact or approximate kernel.
However, numerical implementations of second-order methods
will struggle even when there is no exact kernel
or when the gradient is only partly in it,
and so a numerical index of flatness is required.
This is analogous to the requirement to specify a tolerance
for the norm of the gradient when deciding whether to consider a point
an approximate critical point or not.

We quantify the degree of gradient-flatness of a point
by means of
the \emph{relative residual norm} ($r$)
and the \emph{relative co-kernel residual norm} ($r_H$)
for the Newton update direction $p$.
The vector $p$ is an inexact solution to the Newton system $Hp + g = 0$,
where $H$ and $g$ are the current iterate's Hessian and gradient.
The residual is equal to $Hp + g$,
and the smaller its norm, the better $p$ is as a solution.
The co-kernel residual is equal to the Hessian times the residual,
and so ignores any component in the kernel of the Hessian.
Its norm quantifies the quality of an inexact Newton solution
in the case that the gradient lies partly in the Hessian kernel,
the unsatisfiable case, where $Hp \neq -g$ for any $p$.
When the residual is large but the co-kernel residual is small
(norms near 1 and 0, respectively, following suitable normalization),
then we are at a point where
the gradient is almost entirely in the kernel of the Hessian:
an approximate gradient-flat point.
In the results below, we consider a point approximately gradient-flat
when the value of $r_H$ is below 5e-4
while the value of $r$ is above $0.9$.
See \appref{app:residuals} for definitions and details.
We emphasize that numerical issues for second-order methods
can arise even when the degree of gradient-flatness
is small.

Under this relaxed definition of gradient-flatness,
there will be a neighborhood of approximate gradient-flat points
around a strict, exact gradient-flat point
for functions with Lipschitz-smooth gradients and Hessians.
Furthermore, there might be connected sets of non-null Lebesgue measure
which all satisfy the approximate gradient-flatness condition
but none of which satisfy the exact gradient-flatness condition.
We call both of these \emph{gradient-flat regions}.

There are multiple reasonable numerical indices of flatness besides
the definition above.
For example,
the Hessian-gradient regularity condition in~\cite{roosta2018},
which is used to prove convergence of Newton-MR,
would suggest creating a basis for the approximate kernel of the Hessian
and projecting the gradient onto it.
Alternatively, one could compute the Rayleigh quotient of the gradient with respect to the Hessian.
Our method has the advantage of being computed as part of the Newton-MR algorithm.
It furthermore avoids diagonalizing the Hessian or the specification
of an arbitrary eigenvalue cutoff.
The Rayleigh quotient can be computed with only one Hessian-vector product,
plus several vector-vector products,
so it might be a superior choice for larger problems where
computing a high-quality inexact Newton step is computationally infeasible.

\section{Gradient-Flat Regions are Common on Deep Network Losses}\label{sec:results}

To determine whether gradient-flat regions are
responsible for the poor behavior of Newton methods
on deep neural network (DNN) losses
demonstrated in~\figref{fig:nmr_comparison},
we applied Newton-MR to the loss of a
small, two hidden layer fully-connected autoencoder
trained on 10k MNIST images downsized to 4x4,
similar to the downsized datasets in~\cite{dauphin2014,pennington2017}.
We found similar results on a fully-connected classifier
trained on the same MNIST images via the cross-entropy loss
(see~\appref{app:mlp})
and another classifier trained
on a very small subset of 50 randomly-labeled MNIST images,
as in~\cite{zhang2016}
(see~\appref{app:mem}).
We focused on Newton-MR because we
found that a damped Newton method like that in~\cite{dauphin2014}
performed poorly, as reported for the XOR problem in~\cite{coetzee1997},
and furthermore that there was insufficient detail
to replicate~\cite{dauphin2014} exactly.
We denote the network losses by $L$
and the parameters by $\theta$.
See \appref{app:networks} for details on
the networks and datasets and
\appref{app:cpf} for details on
the critical point-finding experiments.

\begin{figure}[htpb]
    \centering
    \includegraphics[width=0.75\linewidth]{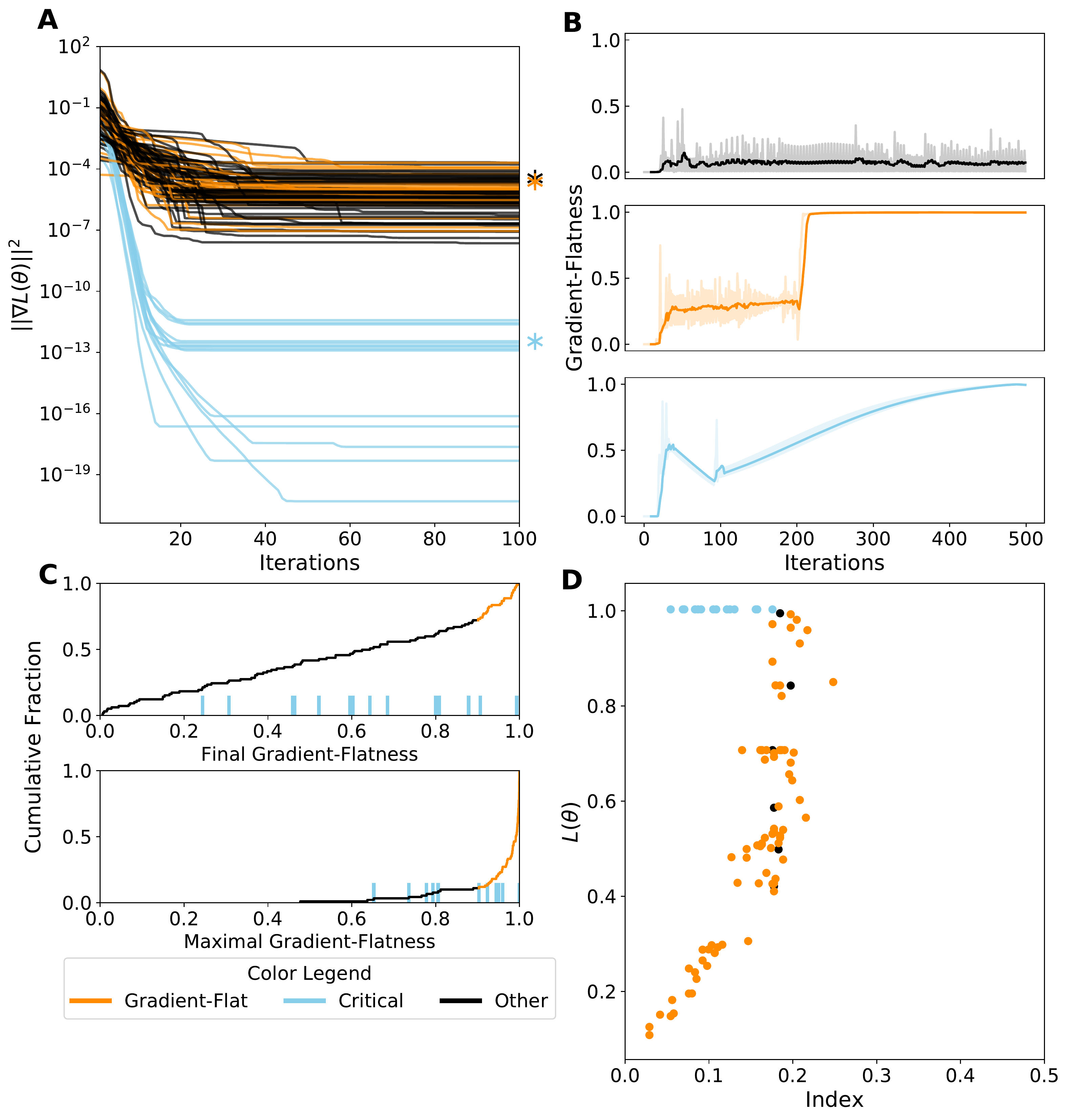}
    \caption{\textbf{Critical Point-Finding Methods 
    More Often Find Gradient-Flat Regions
    on a Neural Network Loss}.
    \textbf{A}:
    Squared gradient norms across the first 100 iterations of Newton-MR
    for 100 separate runs on an auto-encoder loss.
    Gradient norms were flat after 100 iterations.
    See~\appref{app:networks} for details.
    Runs that terminate with squared gradient norm below 1e-10,
    i.e.~at a critical point, in \successcolor{}.
    Runs that terminate above that cutoff and with $r$ above $0.9$,
    i.e.~in a gradient-flat region, in \failcolor{}.
    All other runs in black.
    Asterisks indicate trajectories in $\emph{B}$.
    \textbf{B}:
    The relative residual norm $r$,
    an index of gradient-flatness,
    for the approximate Newton update
    computed by MR-QLP at each iteration
    (solid lines)
    for three representative traces.
    Values are local averages with a window size of 10 iterations.
    Raw values are plotted transparently underneath.
    Top: non-flat, non-critical point (black).
    Middle: flat, non-critical point (\failcolor{}).
    Bottom: flat, critical point (\successcolor{}).
    \textbf{C}:
    Empirical cumulative distribution functions for
    the final (top) and maximal (bottom) relative residual norm $r$ observed
    during each run of Newton-MR.
    Values above the cutoff for approximate gradient-flatness, $r>0.9$,
    in \failcolor{}.
    Observations from runs that terminated below the cutoff for critical points,
    $\sumsq{\grad{L}{\theta}} <$ 1e-10,
    indicated with \successcolor{} ticks.
    \textbf{D}:
    Loss and index for the maximally gradient-flat points
    obtained during application of Newton-MR.
    Points with squared gradient norm below 1e-10 in \successcolor{}.
    Other points colored by their gradient-flatness:
    points above $0.9$ in \failcolor{}, points below in black.
    Only points with squared gradient norm below 1e-4 shown.
    }
    \label{fig:gfp_dnn}
\end{figure}

Gradient norms for the first 100 iterations
appear in \figref{fig:gfp_dnn}A.
As in the non-linear autoencoder applied to the multivariate Gaussian data
(\figref{fig:nmr_comparison}C),
we found that, after 500 iterations,
all of the runs had squared gradient norms
over 10 orders of magnitude greater than the typical
values observed after convergence in the linear case
(<1e-30, \figref{fig:nmr_comparison}A).
14\% of runs terminated with squared gradient norm
below the cutoff in~\cite{frye2019}
and so found likely critical points
(\successcolor{}).
Twice as many runs terminated above that cutoff
but terminated in a gradient-flat region
(28\%, \failcolor{}),
while the remainder were above
the cutoff but were not in a gradient-flat region
at the final iteration
(black).

\afterpage{\clearpage}

The relative residual norm for the Newton solution, $r$,
is an index of gradient-flatness;
see \secref{sec:approxgfp} and \appref{app:residuals} for details.
The values of $r$ for every iteration
of Newton-MR are shown for three representative traces
in \figref{fig:gfp_dnn}B.
In the top trace,
$r$ is close to 0,
indicating that the iterates are not in a gradient-flat region
($r\ll0.9$, black).
Newton methods can be substantially slowed when even a small fraction
of the gradient is in the kernel~\cite{griewank1983}
and can converge to points that are not gradient-flat~\cite{byrd2004}.
By contrast, in the middle trace (\failcolor{}),
the value of $r$ approaches $1$,
indicating that almost the entirety of the gradient is in the kernel.
This run terminated in a gradient-flat region,
at effectively an exactly gradient-flat point.
Further, the squared gradient norm at 500 iterations, 2e-5,
is five orders of magnitude higher than the cutoff,
1e-10.
This is smaller than the minimum observed during optimization
of this loss (squared gradient norms between 1e-4 and 5e1),
indicating the presence of non-critical gradient-flat regions
with very low gradient norm.
Critical point-finding methods that disqualify points
on the basis of their norm will both converge to
and accept these points,
even though they need not be near true critical points.
In the bottom trace (\successcolor{}),
the behavior of $r$ is the same,
while the gradient norm drops much lower, to 3e-13,
suggesting convergence to a gradient-flat region around a critical point
that has an approximately singular Hessian.

Not all traces exhibit such simple behavior for the value of $r$.
In many traces, the value of $r$ oscillates from values close to 1
to middling values,
indicating that the algorithm is bouncing in and out
of one or more gradient-flat regions
(see~\appref{app:mlp} for examples, on a classifier).
This can occur when the final target of convergence
given infinite iterations
is a gradient-flat point,
as in the example in~\secref{sec:toy}.

We found that 99 of 100 traces included a point
where at least half of the gradient was in the kernel,
according to our residual measure,
while 89\% of traces included a point that had
a residual greater than $0.9$,
and 50\% included a point with $r > 0.99$
(\figref{fig:gfp_dnn}C, bottom).
This demonstrates that
there are many regions of substantive gradient-flatness,
in which second-order critical point-finding methods could be substantively slowed.

The original purpose of applying these critical point-finding methods
was to determine whether the no-bad-local-minima property held
for this loss function, and more broadly to characterize
the relationship at the critical points
between the loss and the local curvature,
summarized via the Morse index.
If we look at either the points found after 500 iterations
(results not shown;
see~\appref{app:mlp} for an example on a classifier)
or the iterates with the highest gradient-flatness
(\figref{fig:gfp_dnn}D),
we find that the qualitative features of the loss-index relationship reported in
\cite{dauphin2014} and \cite{pennington2017} are recreated:
convex shape, small spread at low index that increases for higher index,
no minima or near-minima at high values of the loss.
However, our analysis suggests that the majority of these points are not critical points
but either strict gradient-flat points (\failcolor{})
or simply points of spurious or incomplete
Newton convergence (black).
The approximately critical points we do see (\successcolor{})
have a very different loss-index relationship:
their loss is equal to the loss of a network with all parameters set to 0
and their index is low, but not 0.

\section{Discussion}

The results above and in the appendices demonstrate that gradient-flat regions,
where the gradient is nearly in the approximate kernel of the Hessian,
are a prevalent feature of some protoypical neural network loss surfaces
and that many critical point-finding methods are attracted to them.
The networks used in this paper are very small,
relative to practical networks for image recognition
and natural language processing,
which have several orders of magnitude more parameters.
However, increasing parameter count tends to
increase the singularity of loss Hessians%
~\cite{sagun2017},
and so we expect there to be even greater gradient-flatness
for larger networks.

The strategy of using gradient norm cutoffs to determine
whether a point is near enough to a critical point
for the loss and index to match the true value is natural,
but in the absence of guarantees on the smoothness of the behavior of 
the Hessian (and its spectrum) around the critical point,
the numerical value sufficient to guarantee correctness is unclear.
Our observations of gradient-flat regions at extremely low
gradient norm and the separation of these values,
in terms of loss-index relationship,
from the bulk of the observations
suggest that there may be spurious targets
of convergence for critical point-finding methods
even at such low gradient norm.
Alternatively, they may in fact be near real critical points,
and so indicate that the simple, convex picture of loss-index relationship
painted by the numerical results in~\cite{dauphin2014} and~\cite{pennington2017}
is incomplete.
Indeed, recent analytical results
have demonstrated that bad local minima do exist for
almost all neural network architectures and datasets
(see~\cite{ding2019} for a helpful table of positive
and negative theoretical results regarding local minima).
Furthermore, our observation of singular Hessians
at low gradient norm
suggests that some approximate saddle points of neural network losses
may be degenerate (as defined in~\cite{jin2018a})
and non-strict (as defined in~\cite{lee2016}),
which indicates that gradient descent may be attracted to these points,
according to the analyses in~\cite{jin2018a} and~\cite{lee2016}.
These points need not be local minima.
However, in two cases we observe the lowest-index saddles at low values of the loss
(see \figref{fig:gfp_dnn}, \figref{fig:gfp_mlp})
and so these analyses still predict that gradient descent will
successfully reduce the loss,
even if it doesn't find a local minimum.
In the third case,
an over-parameterized network \figref{fig:gfp_mem},
we do observe a bad local minimum,
as predicted in~\cite{ding2019}
for networks capable of achieving 0 training error.

Our results motivate a revisiting of the numerical results
in~\cite{dauphin2014} and~\cite{pennington2017}.
Looking back at Figure 4 of~\cite{dauphin2014},
we see that their non-convex Newton method,
a second-order optimization algorithm designed to avoid saddle points
by reversing the Newton update along directions of negative curvature,
appears to terminate at a gradient norm of order 1.
This is only a single order of magnitude lower than what was observed during training.
It is likely that this point was either in a gradient-flat region
or otherwise had sufficient gradient norm in the Hessian kernel to
slow the progress of their algorithm.
This observation demonstrates that second-order methods
designed for optimization,
which use the loss as a merit function,
rather than norms of the gradient,
can terminate in gradient-flat regions.
In this case, the merit function encourages
convergence to points where the loss,
rather than the gradient norm, is small,
but it still cannot guarantee convergence to a critical point.
The authors of~\cite{dauphin2014} do not report a gradient norm cutoff,
among other details needed to recreate their critical point-finding experiments,
so it is unclear to which kind of points they converged.
If, however, the norms are as large as those of the targets of
their non-convex Newton method,
in accordance with our experience with damped Newton methods
and that of~\cite{coetzee1997},
then the loss-index relationships reported in their Figure 1
are likely to be for gradient-flat points,
rather than critical points.

The authors of~\cite{pennington2017}
do report a squared gradient norm cutoff of 1e-6.
This cutoff is right in the middle of the bulk of values
we observed,
and which we labeled gradient-flat regions and
points of spurious convergence,
based on the cutoff in~\cite{frye2019},
which separates a small fraction of runs from this bulk.
This suggests that some of their putative critical points were gradient flat points.
Their Figure 6 shows a disagreement between their predictions for the index,
based on a loss-weighted mixture of a Wishart and Wigner random matrix,
and their observations.
We speculate that some of this gap is due to their method
recovering approximate gradient-flat points rather than critical points.

Even in the face of results indicating the existence of bad local minima%
~\cite{ding2019},
it remains possible that bad local minima of the loss
are avoided by initialization and optimization strategies.
For example ReLU networks suffer from bad local minima
when one layer's activations are all $0$,
or when the biases are initialized at too small of a value%
~\cite{holzmller2020},
but careful initialization and training can avoid the issue.
Our results do not directly invalidate
this hypothesis,
but they do call the supporting numerical evidence into question.
Our observation of gradient-flat regions on almost every single run
suggests that, while critical points are hard to find
and may even be rare,
regions where gradient norm is extremely small are neither.
For non-smooth losses,
e.g.~those of ReLU networks or networks with max-pooling,
whose loss gradients can have discontinuities,
critical points need not exist,
but gradient-flat regions may.
Indeed, in some cases, the only differentiable minima
in ReLU networks are also flat%
~\cite{laurent2017}.

Other types of critical point-finding methods are not necessarily
attracted to gradient-flat regions, in particular Newton homotopy methods
(first used on neural networks in the 90s~\cite{coetzee1997},
then revived in the 2010s~\cite{ballard2017,mehta2018b}),
which are popular in algebraic geometry~\cite{bates2013}.
However, singular Hessians still cause issues:
for a singular Hessian $H$, the curve to be continued by the homotopy
becomes a manifold with dimension 1 + $\mathrm{rank}\left(H\right)$,
and orientation becomes more difficult.
This can be avoided by removing the singularity of the Hessian,
e.g.~by the randomly-weighted regularization method in~\cite{mehta2018a}.
However, while these techniques may make it possible to find
critical points,
they fundamentally alter the loss surface,
limiting their utility in drawing conclusions about other features
of the loss.
In particular, in the time since
the initial resurgence of interest in the curvature properties
of neural network losses sparked by~\cite{dauphin2014},
the importance of overparameterization for optimization of
and generalization by neural networks has been identified
\cite{li2018,poggio2020}.
Large overparameterized networks have more singular Hessians~\cite{sagun2017},
and so the difference between the original loss and an altered version
with an invertible Hessian is greater.
Furthermore the prevalence of gradient-flat regions should be greater,
since the Hessian kernel covers an increasingly large subspace.

The authors of~\cite{sagun2017} emphasize that
when the Hessian is singular everywhere,
the notion of a basin of attraction is misleading,
since targets of convergence form
connected manifolds
and some assumptions in theorems guaranteeing first-order convergence
become invalid~\cite{jin2018a},
though with sufficient, if unrealistic, over-parameterization
convergence can be proven~\cite{du2018}.
They speculate that a better approach
to understanding the behavior of optimizers
focuses on their exploration of the sub-level sets of the loss.
Our results corroborate that speculation and
further indicate that this flatness means using second-order methods
to try to accelerate exploration of these regions
in search of minimizers
is likely to fail:
the alignment of the gradient with the Hessian's approximate kernel
will tend to produce extremely large steps, for some methods,
or no acceleration and even convergence to non-minimizers,
for others.

Our observation of ubiquitous gradient-flatness further
provides an alternative explanation
for the success and popularity
of approximate second-order optimizers for neural networks,
like K-FAC~\cite{martens2015},
which uses a layerwise approximation to the Hessian.
These methods are typically motivated by appeals to the computational cost
of even Hessian-free exact second-order methods
and their brittleness in the stochastic (non-batch) setting.
However, exact second-order methods are only justified
when the second-order model is good,
and at an exact gradient-flat point,
the second-order model can be infinitely bad,
in a sense, along the direction of the gradient.
Approximations need not share this property.
Even more extreme approximations,
like the diagonal approximations in the adaptive gradient family
(e.g.~AdaGrad~\cite{duchi2011}, Adam~\cite{kingma2014}),
behave very reasonably in gradient-flat regions:
they smoothly scale up the gradient in the directions in which
it is small and changing slowly,
without making a quadratic model that is optimal
in a local sense but poor in a global sense.

Overall, our results underscore the difficulty of searching
for critical points of singular non-convex functions,
including deep network loss functions,
and shed new light on other numerical results in this field.
In this setting, second-order methods for finding critical points can fail badly,
by converging to gradient-flat points.
This failure can be hard to detect unless it is specifically measured.
Furthermore, gradient-flat points are generally places where
quadratic approximations become untrustworthy,
and so our observations are of relevance for the
design of exact and approximate second-order optimization methods as well.

\section*{Acknowledgements}

The authors would like to thank
Yasaman Bahri, Jesse Livezey, Dhagash Mehta, Dylan Paiton, and Ryan Zarcone for useful discussions.
Authors CF \& AL were supported by the National Science Foundation Graduate Research Fellowship Program
under Grant No. DGE 1752814.
NW was supported by the Google PhD Fellowship.
Author AL was supported by a National Institues of Health training grant,
5T32NS095939.
MRD was supported in part by the U. S. Army Research Laboratory and the U. S. Army Research Office
under contract W911NF-13-1-0390.
KB was funded by a DOE/LBNL LDRD, ‘Deep Learning for Science’, (PI, Prabhat).

\bibliography{bibliography}
\bibliographystyle{plain}


\appendix
\renewcommand\thefigure{\thesection.\arabic{figure}}
\setcounter{figure}{0}

\section{Appendices}

\subsection{Networks and Datasets}\label{app:networks}

\subsubsection{Datasets}

For the experiments in \figref{fig:nmr_comparison},
10000 16-dimensional Gaussian vectors with mean parameter 0
and diagonal covariance with linearly-spaced values between 1 and 16
were generated and then mean-centered.

For the experiments in \figref{fig:gfp_dnn} and \figref{fig:gfp_mlp},
10000 images from the MNIST dataset~\cite{lecun2010}
were cropped to 20 x 20 and rescaled to 4 x 4
using PyTorch~\cite{paszke2019},
then $z$-scored.
This was done for two reasons.
Firstly, to improve the conditioning of the data covariance,
which is very poor for MNIST due to low variance in the border pixels.
Secondly, to reduce the number of parameters $n$ in the network,
as computing a high-quality inexact Newton solution is $O(n^2)$.
Non-linear classification networks trained 
on this down-sampled data could still obtain accuracies
above 90\%,
better than the performance of logistic regression ($\approx$87\%).

For the experiments in \figref{fig:gfp_mem},
50 random images of 0s and 1s from the MNIST dataset
were PCA-downsampled to 32 dimensions
using sklearn~\cite{pedregosa2011}.
This provided an alternative approach to
improving the conditioning of the data covariance
and reducing the parameter counts in the network.
The labels for these images were then shuffled.

\subsubsection{Networks}

All networks, their optimization algorithms,
and the critical point-finding algorithms were
defined in the
\texttt{autograd} Python package~\cite{maclaurin2016}.
For the experiments in \figref{fig:nmr_comparison},
two networks were trained:
a linear auto-encoder with
a single, fully-connected hidden layer of 4 units
and a deep non-linear auto-encoder with 
two fully-connected hidden layers of 16 and 4 units
with Swish~\cite{ramachandran2017} activations.
Performance of the critical point-finding algorithms was even worse for networks with rectified linear units
(results not shown)
as reported by others
(Pennington and Bahri, personal communication).
Non-smooth losses need not have gradients that
smoothly approach 0 near local minimizers,
so it is only sensible to apply critical point-finding
to smooth losses, see~\cite{laurent2017}.
The non-linear auto-encoder used
$\ell_2$ regularization.
Neither network had biases.
All auto-encoding networks were trained with
mean squared error.

For the experiments in \figref{fig:gfp_dnn},
a fully-connected autoencoder with two hidden layers
of 8 and 16 units,
with Swish activations and biases, was used.
This network had no $\ell_2$ regularization.

For the experiments in \figref{fig:gfp_mlp},
a fully-connected classifier with two hidden layers
of 12 and 8 units,
with Swish activations and biases, was used.
This network had $\ell_2$ regularization,
since the cross-entropy loss with which it was trained
can otherwise have critical points at infinity.

For the experiments in \figref{fig:gfp_mem},
a fully-connected classifier with two hidden layers
of 32 and 4 units,
with Swish activations, was used.
This network had no biases.
This network also used $\ell_2$ regularization
and was trained with the cross entropy loss.
Networks were trained to near-perfect training performance:
48 to 50 correctly-classified examples out of 50.

\subsection{Critical Point-Finding Experiments}\label{app:cpf}

For all critical point-finding experiments,
we followed the basic procedure pioneered in~\cite{dauphin2014}
and used in~\cite{pennington2017} and~\cite{frye2019}.
First, an optimization algorithm was used to train
the network multiple times.
For the results in~\figref{fig:nmr_comparison},
this algorithm was full-batch gradient descent,
while for the remainder of the results,
this algorithm was full-batch gradient descent with momentum
(learning rates 0.1 in both cases; momentum 0.9 in the latter).

The parameter values produced during optimization
were then used as starting positions for Newton-MR.
Following~\cite{pennington2017} and
based on the arguments in~\cite{frye2019},
we selected these initial points uniformly at random
with respect to their loss value.

Newton-MR~\cite{roosta2018}
computes an inexact Newton update with
the MR-QLP solver~\cite{choi2011}
and then performs back-tracking line search
based on the squared gradient norm.
For pseudocode of the equivalent exact algorithm, see~\appref{app:nmr}

The MR-QLP solver has the following hyperparameters
for determining stopping behavior:
maximum number of iterations (\texttt{maxit}),
maximum solution norm (\texttt{maxxnorm}),
relative residual tolerance (\texttt{rtol}),
condition number limit (\texttt{acondlim}).
We set \texttt{maxit} to be equal
to the number of parameters,
since with exact arithmetic,
this is sufficient to solve the Newton system.
We found that the \texttt{maxxnorm}
and \texttt{acondlim} parameters did not affect
stopping behavior,
which was driven by the tolerance \texttt{rtol}.
For \figref{fig:nmr_comparison},
we used a tolerance of 1e-10.
For \figref{fig:gfp_dnn} and \figref{fig:gfp_mlp},
we used a tolerance of 5e-4,
based on the values for the relative residuals
found after \texttt{maxit} iterations on test points.
See~\appref{app:residuals}
for details about the relative residual
stopping criterion.
We do not provide pseudocode for this algorithm;
see~\cite{choi2011}.

The back-tracking line search has the following hyperparameters:
$\alpha$,
the starting step size,
$\beta$,
the multiplicative factor by which the step size is reduced,
and $\rho$,
the degree of improvement required to terminate the line search.
We set these hyperparameters to
$\alpha\defeq0.1$,
$\beta\defeq0.5$, and
$\rho\defeq0.1$.
Furthermore, in back-tracking line search for Newton methods,
it is important to always check a unit step size
in order to attain supralinear convergence~\cite{nocedal2006}.
So before running the line search,
we also check unit step size with a stricter $\rho^\prime\defeq0.5$.

\subsection{Newton-MR Pseudocode}\label{app:nmr}

\begin{wrapfigure}{L}{0.5\textwidth}
    \begin{minipage}{0.5\textwidth}
        \begin{algorithm}[H]
            \SetAlgoLined{}
            \textbf{Require}
            $T\in \N, \theta_0 \in \R^N,$\ \\
            $\nabla{f}\from\R^N\to\R^N,
            \nabla^2{f}\from\R^N\to\R^{N\times N}\
            $
            \\ \ \\
            $t = 0$\\
            \While{$t < T$}{
                $g \leftarrow \nabla{f}\left(\theta_t\right)$\\
                $H \leftarrow \nabla^2{f}\left(\theta_t\right)$\\
                $P \leftarrow \argmin{p^\prime}{\sumsq{Hp^\prime + g}}$\\
                $p \leftarrow \argminF_{p \in P} \sumsq{p}$\\
                $\alpha \leftarrow \argmin{\alpha}{\sumsq{\grad{f}{\theta_t + \alpha p}}}$\\
                $\theta_{t-1} \leftarrow \theta_t + \alpha p$\\
                \If{$\theta_t == \theta_t$}{\textbf{break}}
                $t \leftarrow t + 1$
            }
            \label{algo:nmr}
            \caption{Exact Newton-MR}
      \end{algorithm}
    \end{minipage}
\end{wrapfigure}

The pseudocode in \algoref{algo:nmr} defines an exact least-squares Newton method
with exact line search.
To obtain the inexact Newton-MR algorithm,
the double $\argminF$ to determine the update direction $p$
should be approximately satisfied using the MR-QLP solver%
~\cite{choi2011}
and the $\argminF$ to determine the step size $\alpha$
should be approximately satisfied using
Armijo-type back-tracking line search.
For details, see~\cite{roosta2018}.

\subsection{Relative Residual and Relative Co-Kernel Residual}\label{app:residuals}

The relative residual norm, $r$,
measures the size of the error
of an approximate solution to the Newton system.
Introducing the symbols $H$ and $g$,
for the Hessian and gradient at a query point,
the Newton system may be written
$0 = Hp + g$
and $r$ is then
$$
    r(p) = \frac{\norm{Hp + g}}{\norm{H}_F \norm{p} + \norm{g}}
$$
where $\norm{M}_F$ of a matrix $M$ is its Frobenius norm.
Since all quantities are non-negative,
$r$ is non-negative;
because the denominator bounds the numerator,
by the triangle inequality and the compatibility
of the Frobenius and Euclidean norms,
$r$ is at most 1.
For an exact solution of the Newton system $\star{p}$,
$r(\star{p})$ is 0, the minimum value,
while $r(0)$ is 1, the maximum value.
Note that small values of $\norm{p}$
do not imply large values of this quantity,
since $\norm{p}$ goes to 0 when a Newton method
converges towards a critical point,
while $r$ goes to 0.

When $g$ is partially in the kernel of $H$,
the Newton system is unsatisfiable,
as $g$ will also be partly in the co-image of $H$,
the linear subspace into which $H$ cannot map any vector.
In this case, the minimal value for $r$ will no longer be $0$.
The optimal solution for $\norm{Hp + g}$
instead has the property that its residual
is $0$ once restricted to the co-kernel of $H$,
the linear subspace orthogonal to the kernel of $H$.
This co-kernel residual can be measured by applying the matrix $H$
to the residual vector $Hp + g$.
After normalization, it becomes
$$
    r_H(p) = \frac{\norm{H(Hp + g)}}{\norm{H}_F \norm{Hp + g}}
$$
Note that this value is also small when the gradient lies
primarily along the eigenvalues of smallest magnitude
On each internal iteration, MR-QLP checks whether
either of these values is below a tolerance level --
in our experiments, 5e-4 --
and if either is, it ceases iteration.
With exact arithmetic,
either one or the other of these values
should go to 0 within a finite number of iterations;
with inexact arithmetic, they should just become small.
See \cite{choi2011} for details.
Less than 5\% of Newton steps
were obtained from the kernel residual
going below the tolerance,
indicating that almost all points of the loss surface
had an approximately unsatisfiable Newton system.
\clearpage
\subsection{Replication of Results from \secref{sec:results} on MNIST MLP}\label{app:mlp}

\begin{figure}[!h]
    \centering
    \includegraphics[width=0.75\linewidth]{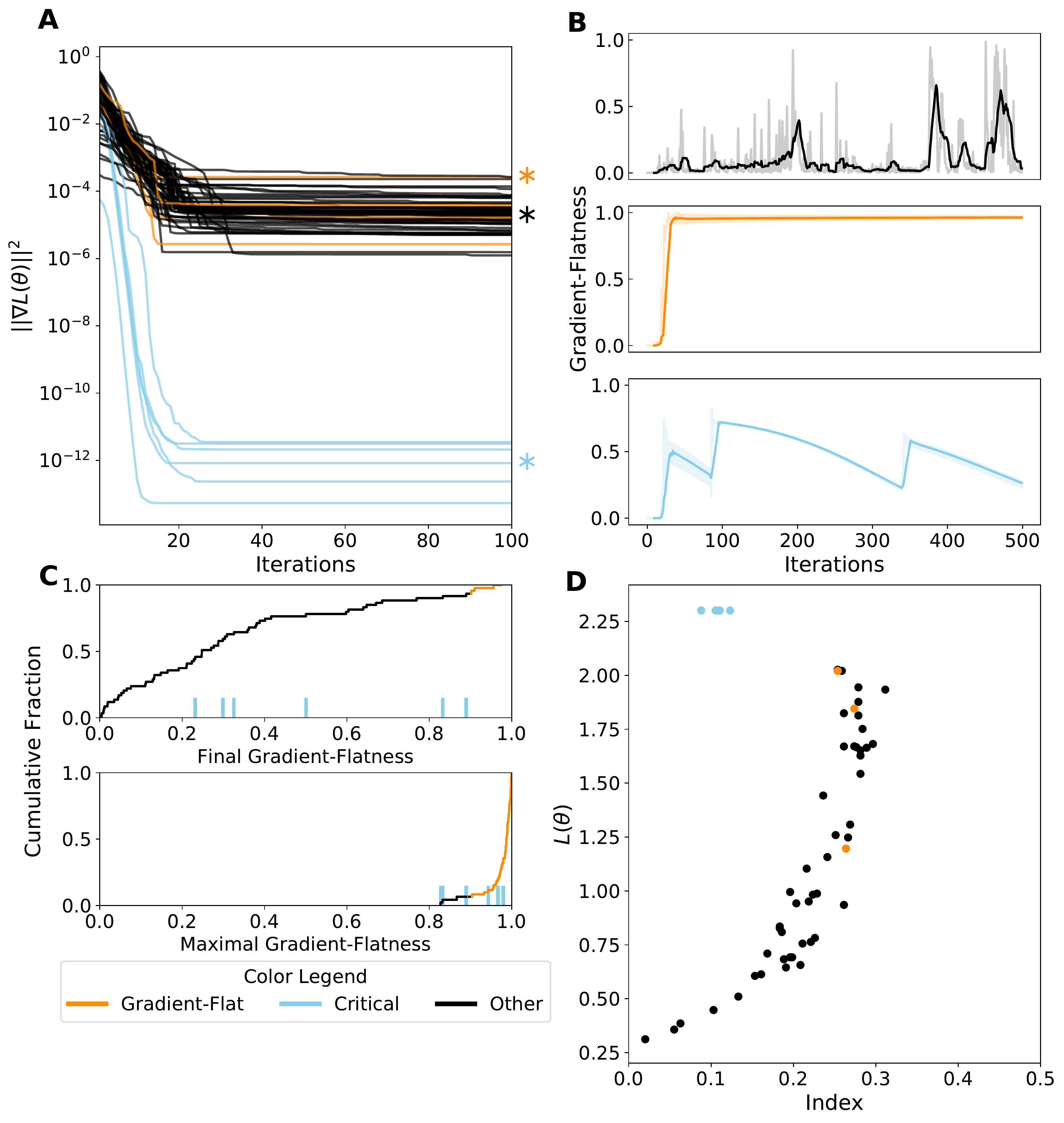}
    \caption{\textbf{Gradient-Flat Regions Also Appear on an MLP Loss}.
    \textbf{A}:
    Squared gradient norms across the first 100 iterations of Newton-MR
    for 60 separate runs on an MLP loss
    (see~\appref{app:networks} for details).
    Runs that terminate with squared gradient norm below 1e-10
    in \successcolor{}.
    Runs that terminate above that cutoff and with $r$ above $0.9$,
    in \failcolor{}.
    All other runs in black.
    Asterisks indicate trajectories in $\emph{B}$.
    \textbf{B}:
    The relative residual norm $r$,
    for the approximate Newton update
    computed by MR-QLP at each iteration
    for three representative traces.
    Values are local averages with a window size of 10 iterations.
    Raw values are plotted transparently underneath.
    Top: non-flat, non-critical point (black).
    Middle: flat, non-critical point (\failcolor{}).
    Bottom: non-flat, critical point (\successcolor{}).
    \textbf{C}:
    Empirical cumulative distribution functions for
    the final (top) and maximal (bottom) relative residual norm $r$.
    Values above the cutoff for approximate gradient-flatness, $r>0.9$,
    in \failcolor{}.
    Observations from runs that terminated below the cutoff for critical points,
    $\sumsq{\grad{L}{\theta}} <$ 1e-10,
    indicated with \successcolor{} ticks.
    \textbf{D}:
    Loss and index for the points found
    after 500 iterations of Newton-MR.
    Colors as in top-left; only points with squared gradient norm below 1e-4 shown.}
    \label{fig:gfp_mlp}
\end{figure}

We repeated the experiments whose results are visualized in~\figref{fig:gfp_dnn}
on the loss surface of a fully-connected classifier on the same modified MNIST dataset
(details in~\appref{app:networks}).
We again found that the performance of
the Newton-MR critical point-finding algorithm was poor
(\figref{fig:gfp_mlp}A)
and that around 90\% of runs encountered a point
with gradient-flatness above $0.9$
(\figref{fig:gfp_mlp}C, bottom row).
However, we observed that fewer runs terminated
at a gradient-flat point
(\figref{fig:gfp_mlp}C, top row),
perhaps because the algorithm was bouncing in and out
of gradient-flat regions (\figref{fig:gfp_mlp}B, top and bottom rows),
rather than because of another type of spurious Newton convergence.
If we measure the loss-index relationship at these non-gradient-flat points
(\figref{fig:gfp_mlp}D),
we see the same pattern as in~\figref{fig:gfp_dnn}.
This also holds if we look at the maximally flat points,
as in~\figref{fig:gfp_dnn}D (results not shown).

\subsection{Replication of Results from \secref{sec:results} on Binary MNIST Subset Memorization}\label{app:mem}

\begin{figure}[!h]
    \centering
    \includegraphics[width=0.75\linewidth]{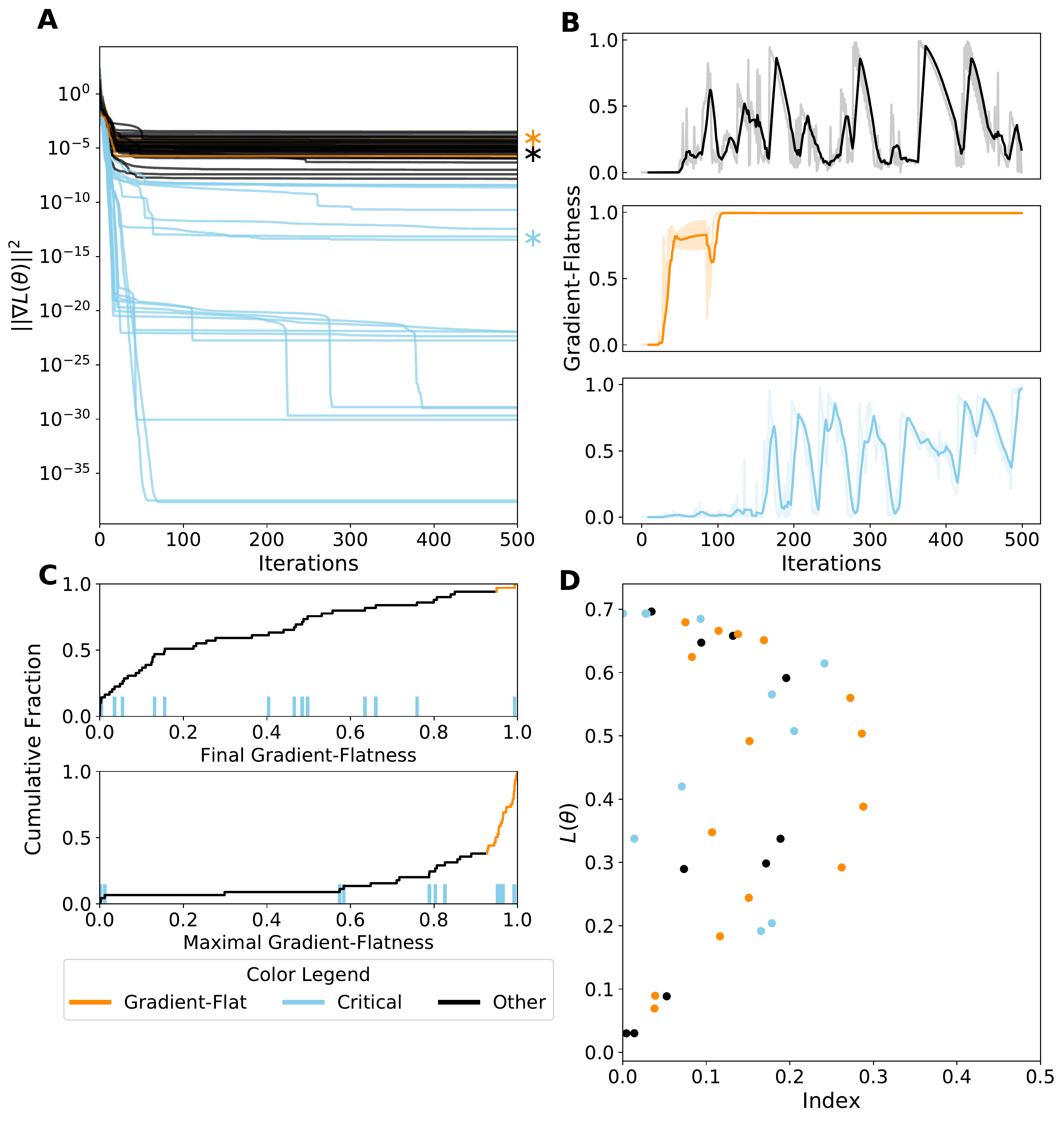}
    \caption{\textbf{Gradient-Flat Regions Also Appear on an Over-Parameterized Loss}.
    \textbf{A}:
    Squared gradient norms across 500 iterations of Newton-MR
    for 50 separate runs on the loss of an over-parameterized network
    (see~\appref{app:networks} for details).
    Runs that terminate with squared gradient norm below 1e-10
    in \successcolor{}.
    Runs that terminate above that cutoff and with $r$ above $0.9$,
    in \failcolor{}.
    All other runs in black.
    Asterisks indicate trajectories in $\emph{B}$.
    \textbf{B}:
    The relative residual norm $r$,
    for the approximate Newton update
    computed by MR-QLP at each iteration
    for three representative traces.
    Values are local averages with a window size of 10 iterations.
    Raw values are plotted transparently underneath.
    Top: non-flat, non-critical point (black).
    Middle: flat, non-critical point (\failcolor{}).
    Bottom: flat, critical point (\successcolor{}).
    \textbf{C}:
    Empirical cumulative distribution functions for
    the final (top) and maximal (bottom) relative residual norm $r$.
    Values above the cutoff for approximate gradient-flatness, $r>0.9$,
    in \failcolor{}.
    Observations from runs that terminated below the cutoff for critical points,
    $\sumsq{\grad{L}{\theta}} <$ 1e-10,
    indicated with \successcolor{} ticks.
    \textbf{D}:
    Loss and index for the points found
    after 500 iterations of Newton-MR.
    Colors as in top-left; only points with squared gradient norm below 1e-4 shown.}
    \label{fig:gfp_mem}
\end{figure}

We repeated the experiments whose results are visualized in~\figref{fig:gfp_dnn}
on the loss surface of a fully-connected classifier on a small subset
of 50 0s and 1s from the MNIST dataset
(details in~\appref{app:networks}).
In this setting, the network is over-parameterized,
in that it has a hidden layer almost as wide as the number of points in the dataset (32 vs 50)
and has more parameters than there are points in the dataset (1160 vs 50).
It is also capable of achieving 100\% accuracy on this training set,
which has random labels, as in~\cite{zhang2016}.
We again observe that the majority of runs
do not terminate with squared gradient norm under 1e-8
(33 out of 50 runs)
and a similar fraction
(31 out of 50 runs)
encounter gradient-flat points
(\figref{fig:gfp_mem}A and C, bottom panel).
The loss-index relationship looks qualitatively different,
as might be expected for a task with random labels.
Notice the appearance of a bad local minimum:
the \successcolor{} point at index 0 and loss $\ln(2)$.

\end{document}